\documentclass[11pt]{article}

\usepackage[final]{acl}

\usepackage{times}
\usepackage{latexsym}

\usepackage[T1]{fontenc}

\usepackage[utf8]{inputenc}

\usepackage{microtype}

\usepackage{inconsolata}

\usepackage{graphicx}

\usepackage{amssymb}
\usepackage{amsmath}
\usepackage{subcaption}
\usepackage{booktabs}
\usepackage{tabularx}

\title{Multilingual Safety Signals Are Multi-Layered: Filtering Safety-Degrading Data for Safer LLMs}

\author{
\textbf{Jiakun Li\textsuperscript{1,2}},
\textbf{Guowei Song\textsuperscript{1}},
\textbf{Sijia Li\textsuperscript{2}},
\textbf{Xingwei He\textsuperscript{3}},
\textbf{Hongzheng Chai\textsuperscript{1}},
\textbf{Yuan Yuan\textsuperscript{1,2,3,4}\thanks{Corresponding author.}} \\
\textsuperscript{1}School of Computer Science and Engineering, Beihang University \\
\textsuperscript{2}Zhongguancun Laboratory \\
\textsuperscript{3}Hangzhou Innovation Institute, Beihang University \\
\textsuperscript{4}Qingdao Research Institute, Beihang University \\
\texttt{lijiakun25@buaa.edu.cn}, \texttt{yuan21@buaa.edu.cn} \\
}

\begin{document}
\maketitle
\begin{abstract}
Preserving safety alignment during large language models fine-tuning is critical, however, recent studies have demonstrated that even benign fine-tuning data may contain safety-degrading samples that silently undermine safety alignment. 
Existing approaches typically identify such samples using representations from a single safety-sensitive layer.
While this assumption has shown effectiveness in monolingual settings, its validity for multilingual models remains unclear due to potential cross-lingual differences in representation patterns.
Through a cross-lingual analysis, we show that sensitive layers are only partially shared across languages, with safety-relevant signals often distributed across multiple layers.
Motivated by these observations, we propose \textsc{MMSafe}, a multi-layer framework for multilingual safety-degrading data identification that captures both shared and language-specific safety signals.
Extensive experiments across multiple models, languages, and safety benchmarks demonstrate that \textsc{MMSAFE} reduces the average harmful-response ratio by 60\% compared with random filtering and achieves stronger average performance than the strongest single-layer baseline, demonstrating the effectiveness of multi-layer modeling for robust multilingual safety alignment. \footnote{We release code at \url{https://github.com/jiakunli-1/MMSafe} to facilitate future research.}

\end{abstract}
\begin{figure*}[t]
  \centering
  \includegraphics[width=\textwidth]{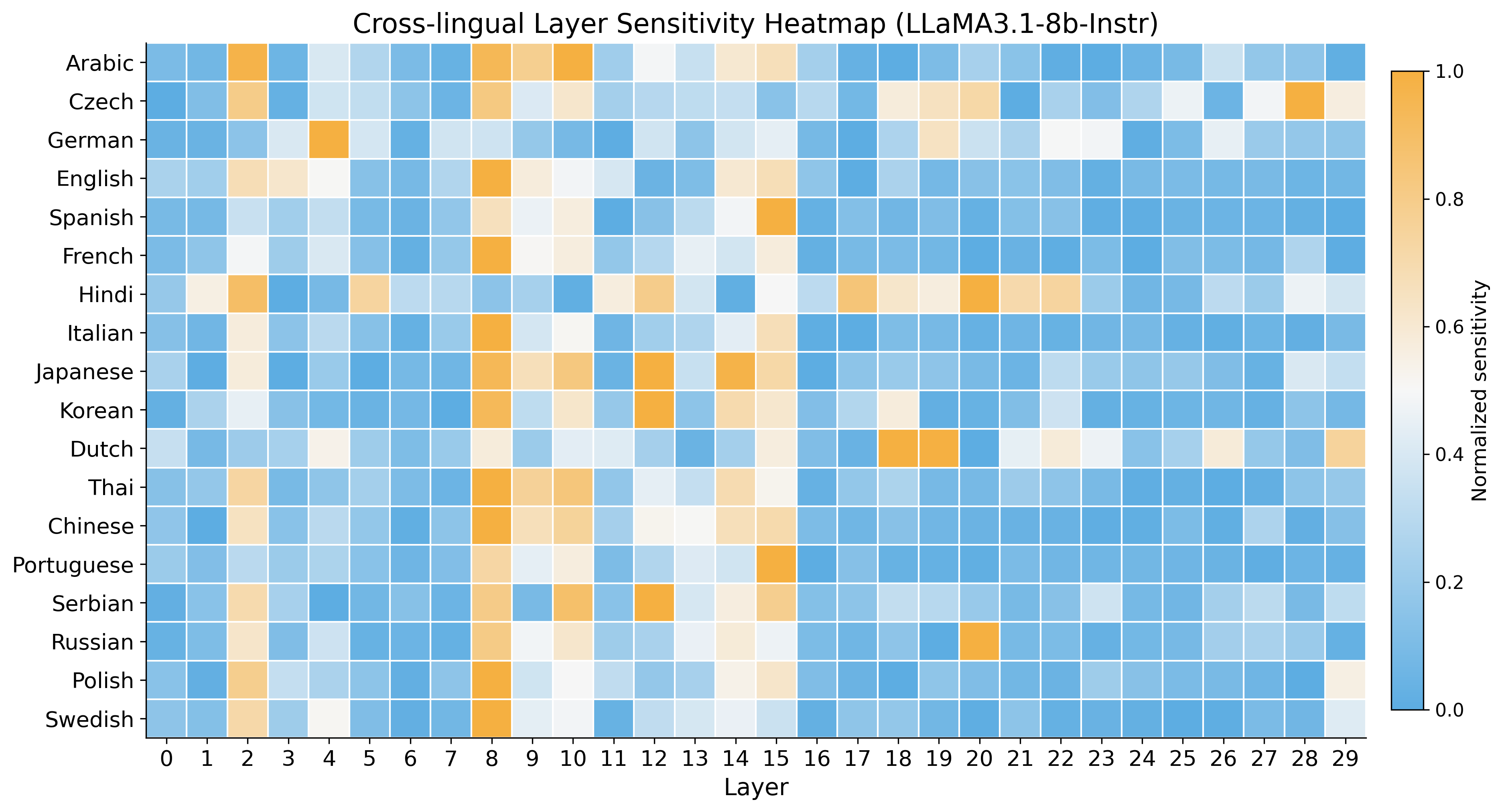}
  \vspace{-0.8em}
  \caption{
Cross-lingual layer sensitivity of LLaMA3.1-8B-Instruct across 18 languages.
Each row is normalized within a language, and darker orange indicates higher sensitivity.
The heatmap reveals a mixed pattern: some layers are consistently salient across languages, while others form language- or language-group-specific sensitivity peaks.
}
  \vspace{-1.2em}
  \label{fig:LLAMA3_1_sensitivity_heatmap}
\end{figure*}

\section{Introduction}

Large language models (LLMs) are commonly aligned for safety before being adapted to downstream tasks through fine-tuning \citep{Ouyang0JAWMZASR22}. 
While fine-tuning is essential for improving task performance and adapting models to downstream domains, recent studies show that its safety outcomes can be strongly influenced by the composition of the training data \citep{Qi0XC0M024, heyour2024}. 
In particular, seemingly benign fine-tuning corpora may contain safety-degrading samples that weaken built-in safeguards and increase the risk of unsafe behaviors after adaptation \citep{bach2026continual}. 
Consequently, identifying and filtering such samples has become a critical step toward preserving safety alignment in practical LLM fine-tuning \citep{DBLP:conf/nips/HuangHIT024, DBLP:conf/iclr/ShenCDC25}.

Recent studies have begun to identify safety-degrading samples through the internal representations of aligned LLMs \citep{heyour2024, DBLP:conf/emnlp/LiLLWLSS25}, suggesting that layer-level activations capture informative safety-relevant signals \citep{chen2026towards, wang2026safeneuron}. 
A representative approach is LARF, which identifies a safety-sensitive layer and leverages its hidden representations to distinguish safety-degrading samples \citep{DBLP:conf/emnlp/LiLLWLSS25}.
As LLMs are increasingly deployed across languages, it becomes important to understand whether such representation-based filtering remains reliable beyond monolingual settings.
However, existing methods largely rely on representations from a single safety-sensitive layer, leaving it unclear whether such a single-layer view remains sufficient in multilingual settings, where safety-relevant signals may not be consistently localized across languages \citep{bu2026align}.

To better understand this limitation, we conduct a cross-lingual analysis of layer sensitivity. 
We find that some sensitive layers are consistently salient across languages, forming language-shared safety signals, while others are salient only for specific languages or language groups, reflecting language-specific safety signals. 
Moreover, multiple layers exhibit comparable sensitivity and capture semantically different safety-related features, suggesting that safety-relevant signals are not uniquely concentrated in a single layer. 
These observations indicate that a single-layer view may be insufficient for multilingual safety-degrading data identification.

Motivated by these observations, we propose \textsc{MMSafe} (\textbf{M}ultilingual \textbf{M}ulti-layer \textbf{Safe}ty Filter), a multi-layer framework for multilingual safety-degrading data identification. 
Instead of relying on a single safety-sensitive layer, \textsc{MMSafe} jointly leverages shared layers that capture cross-lingually consistent safety signals and language-specific layers that reflect language-dependent variation. 
By jointly modeling these complementary signals, \textsc{MMSafe} captures multilingual safety-relevant behavior more comprehensively and reduces the risk of missing safety-degrading samples that are only salient in specific languages or layers.

We conduct extensive experiments across four instruction-tuned LLMs, two multilingual fine-tuning data pools, and three multilingual safety benchmarks. 
The results show that \textsc{MMSafe} consistently reduces harmful responses after fine-tuning. 
On average, \textsc{MMSAFE} reduces the harmful-response ratio by 60.7\% compared with random filtering and achieves a 6.0 percentage-point improvement over the strongest single-layer baseline, outperforming it on three of four evaluated models, with minimal degradation in model utility.

Our main contributions are summarized as follows:
\begin{itemize}
    \item We conduct a cross-lingual layer sensitivity analysis and show that sensitive layers are only partially shared across languages, while safety-relevant signals are distributed across multiple semantically complementary layers.
    
    \item We propose \textsc{MMSafe}, a multilingual multi-layer safety filter for identifying safety-degrading samples. 
    \textsc{MMSafe} jointly leverages language-shared and language-specific sensitive layers, and uses safe-unsafe contrastive representations to score candidate fine-tuning samples.
    
    \item We evaluate \textsc{MMSafe} across multiple models, fine-tuning datasets, languages, and safety benchmarks. 
    Experimental results show that MMSAFE substantially reduces harmful responses compared with random filtering and achieves stronger or comparable performance than strong baselines across model families.
\end{itemize}


\section{Motivation}

\subsection{Questioning the Single-Layer Assumption}

Existing methods rely on a single safety-sensitive layer for identifying safety-degrading samples.
This design implicitly assumes that one layer can provide a sufficiently informative view for distinguishing safety-degrading samples.

In multilingual settings, this assumption becomes questionable for two reasons.
First, layer sensitivity may vary across languages, such that a layer informative for one language may not generalize to another. Second, safety-relevant signals may be distributed across multiple layers, meaning that any single layer captures only a partial view.
Such omissions can be more pronounced in multilingual scenarios, where safety signals may vary across languages as well as across layers.
These considerations motivate our analysis of how safety-relevant signals vary across both languages and layers.

\subsection{Sensitive Layers Are Partially Shared Across Languages}

We first examine whether layer sensitivity is language-dependent.
Following LARF \citep{DBLP:conf/emnlp/LiLLWLSS25}, we estimate
layer-wise sensitivity by scaling the attention and feed-forward
submodules of each layer upward and downward while keeping all other
layers fixed, and measuring the resulting difference in safety-response
counts on multilingual harmful queries.
This sensitivity measure is formally defined in
Section~\ref{sec:multilingual-sensitive-layer-discovery}.
We conduct the analysis on a multilingual harmful query set constructed
from OR-Bench \citep{DBLP:conf/icml/CuiCSH25} and PolyGuardMix
\citep{kumar2025polyguard}, covering 18 languages and 36K queries in total.

The heatmap reveals a clear partially shared structure: while Layer 8 is consistently salient across many languages, other peaks are language-specific, such as Layers 12--14 for Japanese and Korean and Layers 18--19 and 29 for Dutch. 
This indicates that layer sensitivity is only partially shared across languages: some layers are consistently salient, while others are language-specific, meaning that a useful layer for one language may not be equally useful for another.
Similar patterns are observed on LLaMA3-8B-Instruct, Qwen3-8B, and Qwen3-32B, as shown in Appendix~\ref{app:sensitive-further-analysis}.
This partial sharing suggests that no single layer suffices, which raises a further question: do multiple sensitive layers capture redundant or complementary information?

\subsection{Safety Signals Span Multiple Layers}

\begin{figure}[t]
  \centering
  \includegraphics[width=\linewidth]{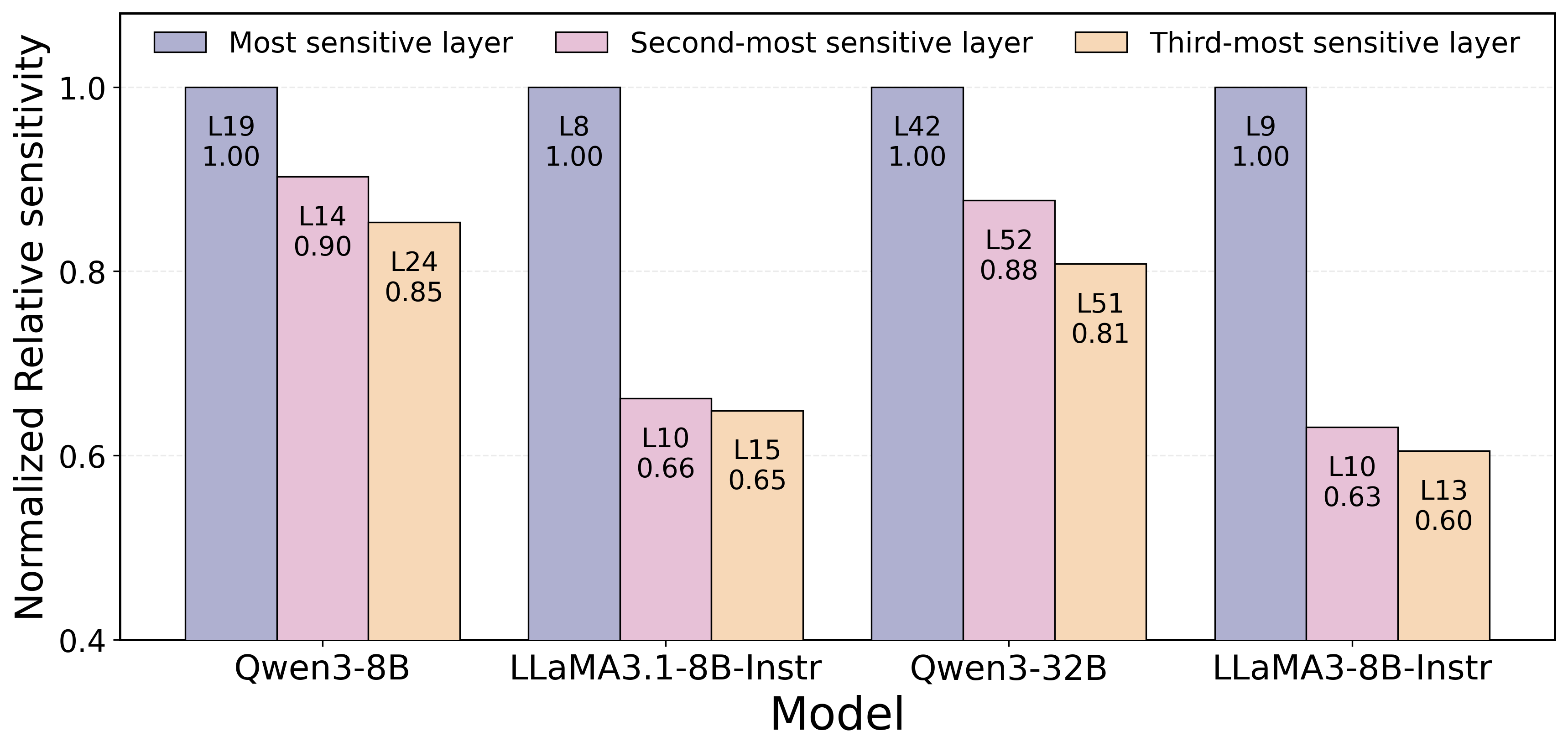}
  \vspace{-1.2em}
  \caption{Relative sensitivity of the top-3 sensitive layers across models.
Sensitivity is normalized by the most sensitive layer within each model.
}
  \vspace{-1.2em}
  \label{fig:top3_sensitive_layers}
\end{figure}

We further examine whether safety-relevant signals are uniquely concentrated in the most sensitive layer. 
Figure~\ref{fig:top3_sensitive_layers} compares the relative sensitivity of the top-3 sensitive layers across models, with each model normalized by its most sensitive layer. 
The second- and third-ranked layers retain substantial relative sensitivity in all models, indicating that the top-ranked layer is often not uniquely dominant. 
This suggests that safety-relevant signals may span multiple sensitive layers rather than being concentrated in a single layer.

To further interpret what each layer captures, we inspect features extracted by sparse autoencoders (SAEs), which decompose hidden representations into sparse and interpretable semantic features.
Specifically, we analyze SAE-interpreted features \citep{arditi2025misalignedpersona} activated by HarmBench \citep{DBLP:conf/icml/MazeikaPYZ0MSLB24} harmful queries on LLaMA3.1-8B-Instruct.
As shown in Table~\ref{tab:sae-layer-concepts}, different layers correspond to different types of harmful concepts. 
Layer 3 mainly captures lexical or local semantic cues, while Layer 7 is more associated with concrete harmful scenarios. 
Layer 11 focuses more on harmful intent and deception, whereas Layer 15 captures broader safety-policy violations. 
These observations suggest that different layers provide semantically complementary safety information.

Together, the sensitivity comparison and SAE feature analysis indicate that safety-relevant signals are both distributed and complementary across layers. 
These findings jointly motivate a design that leverages both language-shared and language-specific sensitive layers, and exploits their complementary safety signals rather than relying on any single layer.
Additional SAE-based analyses of how different layers respond to different harmful categories are provided in Appendix~\ref{app:safety-signals-analysis}.

\begin{table}[t]
\centering
\small
\setlength{\tabcolsep}{3pt}
\renewcommand{\arraystretch}{1.12}
\begin{tabularx}{\linewidth}{c p{0.28\linewidth} X}
\toprule
Layer & Main focus & Representative concepts \\
\midrule
3 
& Lexical / local semantic cues
& Spoofing; lying; spying \\

7 
& Concrete harmful scenarios
& Hoaxes; crimes; dangerous activities; bomb/gun making \\

11 
& Harmful intent and deception
& Fraud; illicit requests; dangerous instructions; hiding evidence \\

15 
& Broad safety-policy violations
& Illegal activities; abuse/hate; evasion; conspiracy theories \\
\bottomrule
\end{tabularx}
\caption{
Representative SAE-interpreted features activated by HarmBench harmful queries on LLaMA3.1-8B-Instruct.
}
\label{tab:sae-layer-concepts}
\vspace{-0.7em}
\end{table}

\begin{figure*}[t]
  \centering
  \includegraphics[width=\textwidth]{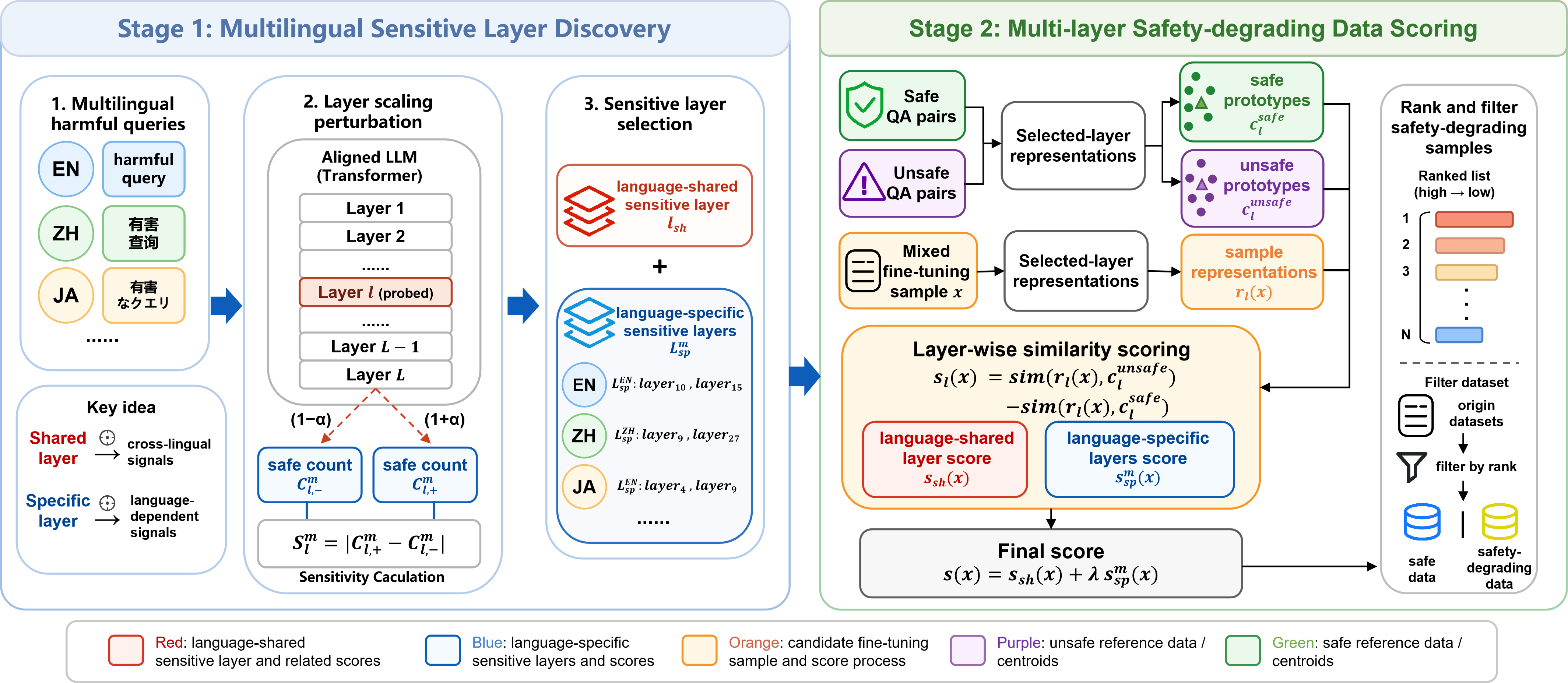}
  \vspace{-0.8em}
  \caption{
  Overview of \textsc{MMSafe}. 
  The framework first discovers language-shared and language-specific sensitive layers through multilingual layer perturbation, and then uses safe-unsafe contrastive representations from these layers to score and filter safety-degrading samples.
  }
  \vspace{-1.0em}
  \label{fig:method-framework}
\end{figure*}

\section{Method}
Motivated by our finding that safety-relevant signals are distributed across both shared and language-specific layers, we propose \textsc{MMSafe}, a multilingual multi-layer safety filter for identifying safety-degrading samples in multilingual fine-tuning data.
\textsc{MMSafe} consists of two stages.
First, we discover sensitive layers by perturbing each layer and measuring changes in safety response counts on multilingual harmful queries, yielding both \emph{language-shared sensitive layers} and \emph{language-specific sensitive layers}. 
Second, we construct safe-unsafe contrastive representations from these layers and aggregate their signals to score and filter safety-degrading samples.
Figure~\ref{fig:method-framework} provides an overview of this two-stage framework.

\subsection{Multilingual Sensitive Layer Discovery}
\label{sec:multilingual-sensitive-layer-discovery}

We first identify safety-sensitive layers in multilingual settings. Let $\mathcal{M}$ denote the set of languages and let the aligned LLM contain $L$ transformer layers. For each language $m \in \mathcal{M}$, we use a set of harmful queries $\mathcal{D}^{m}_{\mathrm{harm}}$ to estimate the safety sensitivity of each layer under controlled perturbations.

For each transformer layer $l \in \{0,\ldots,L-1\}$, we apply upward and downward scaling perturbations to the layer submodules. Let $A_l$ and $F_l$ denote the attention and feed-forward submodules of layer $l$, respectively. Given a perturbation strength $\alpha \in (0,1)$, we define:
\begin{equation}
\label{eq:scaled-modules}
A_l^{\pm} = (1 \pm \alpha) A_l,
\qquad
F_l^{\pm} = (1 \pm \alpha) F_l .
\end{equation}
When evaluating layer $l$, all other layers remain unchanged, and only $A_l$ and $F_l$ are replaced by their scaled versions. Larger behavioral shifts indicate stronger safety relevance of the perturbed layer.

For each scaling direction $\delta \in \{-,+\}$, let
$y_{l,\delta}^{m}(x)=\mathrm{LLM}(x; A_l^{\delta}, F_l^{\delta})$
denote the response generated by the scaled model for query $x$. 
We define the safety response count as:
\vspace{-0.4em}
\begin{equation}
\label{eq:safety-response-count}
C_{l,\delta}^{m}
=
\sum_{x \in \mathcal{D}_{\mathrm{harm}}^{m}}
\mathbb{I}_{\mathrm{safe}}
\bigl(
y_{l,\delta}^{m}(x)
\bigr),
\vspace{-0.4em}
\end{equation}
where $\mathbb{I}_{\mathrm{safe}}(\cdot)$ returns $1$ if the response is classified as safe, e.g., refusing to provide harmful instructions, and $0$ otherwise.

The layer sensitivity score $S_l^m$ is defined as the absolute difference in safety response counts between the two scaling directions:
\vspace{-0.4em}
\begin{equation}
\label{eq:layer-sensitivity}
S_l^m =
\left|
C_{l,+}^{m}
-
C_{l,-}^{m}
\right|.
\vspace{-0.4em}
\end{equation}

After computing $S_l^m$, we rank layers for each language according to their sensitivity scores. Let $\mathcal{T}^{m}$ denote the ranked list of sensitive layers for language $m$. We identify the language-shared sensitive layer as the layer with the highest aggregated sensitivity across languages:
\vspace{-0.3em}
\begin{equation}
\label{eq:shared-layer}
l_{\mathrm{sh}}
=
\arg\max_{l}
\sum_{m \in \mathcal{M}} S_l^m .
\vspace{-0.3em}
\end{equation}

For each language $m$, we then select the top-$k$ layers from $\mathcal{T}^{m}$ after excluding the shared layer:
\begin{equation}
\label{eq:specific-layer-set}
\mathcal{L}_{\mathrm{sp}}^{m}
=
\mathrm{Top}_{k}
\left(
\mathcal{T}^{m} \setminus \{l_{\mathrm{sh}}\}
\right).
\end{equation}
This design enables MMSAFE to capture both universally shared safety representations and language-dependent safety characteristics.

\subsection{Multi-layer Safety-degrading Data Scoring}

After identifying the shared and language-specific sensitive layers, \textsc{MMSafe} uses them to score candidate fine-tuning samples.
We first prepare a small reference set of safe and unsafe question-answer pairs, denoted as $\mathcal{D}_{\mathrm{safe}}$ and $\mathcal{D}_{\mathrm{unsafe}}$, respectively. 
For a question-answer pair $x=(q,a)$, we feed the concatenated sequence into the aligned LLM and extract the hidden states at layer $l$. 
Let $\mathbf{H}_l(x) \in \mathbb{R}^{n \times d}$ denote the layer-$l$ hidden states, where $n$ is the sequence length and $d$ is the hidden dimension. 
The sample representation is obtained via pooling over layer-l hidden states:
\vspace{-0.3em}
\begin{equation}
\label{eq:layer-representation}
\mathbf{r}_{l}(x)
=
\mathrm{Pool}
\left(
\mathbf{H}_l(x)
\right).
\vspace{-0.3em}
\end{equation}
In our experiments, $\mathrm{Pool}(\cdot)$ denotes mean pooling over response tokens. For each selected layer $l$, we extract the hidden representation of each reference example and construct safe and unsafe layer prototypes:
\vspace{-0.7em}
\begin{equation}
\label{eq:safe-prototype}
\mathbf{c}_{l}^{\mathrm{safe}}
=
\frac{1}{|\mathcal{D}_{\mathrm{safe}}|}
\sum_{x \in \mathcal{D}_{\mathrm{safe}}}
\mathbf{r}_{l}(x),
\end{equation}
\vspace{-0.7em}
\begin{equation}
\label{eq:unsafe-prototype}
\mathbf{c}_{l}^{\mathrm{unsafe}}
=
\frac{1}{|\mathcal{D}_{\mathrm{unsafe}}|}
\sum_{x \in \mathcal{D}_{\mathrm{unsafe}}}
\mathbf{r}_{l}(x),
\end{equation}
where $\mathbf{r}_{l}(x)$ denotes the representation of sample $x$ at layer $l$.

For a candidate sample $x$ written in language $m$, we extract its representation $\mathbf{r}_{l}(x)$ from each selected layer. 
For languages covered by sensitive-layer discovery, we use both the
language-shared and corresponding language-specific sensitive layers.
For candidate samples in other languages, we use only the
language-shared layer, which provides a language-agnostic fallback.
We measure the similarity between the candidate representation and the safe/unsafe prototypes:
\begin{equation}
\label{eq:unsafe-similarity}
s_l^{\mathrm{unsafe}}(x)
=
\cos\bigl(
\mathbf{r}_{l}(x),
\mathbf{c}_{l}^{\mathrm{unsafe}}
\bigr).
\end{equation}
\vspace{-0.6em}
\begin{equation}
\label{eq:safe-similarity}
s_l^{\mathrm{safe}}(x)
=
\cos\bigl(
\mathbf{r}_{l}(x),
\mathbf{c}_{l}^{\mathrm{safe}}
\bigr).
\end{equation}

The layer-level safety-degrading score is then defined as:
\begin{equation}
\label{eq:layer-score}
s_l(x)
=
s_l^{\mathrm{unsafe}}(x)
-
s_l^{\mathrm{safe}}(x).
\end{equation}
A larger $s_l(x)$ indicates that the sample is closer to unsafe behavior than safe behavior at layer $l$.

We then compute two scores: one from the language-shared sensitive layer and the other from language-specific sensitive layers. 
The shared-layer score is:
\vspace{-0.2em}
\begin{equation}
\label{eq:shared-score}
s_{\mathrm{sh}}(x)
=
s_{l_{\mathrm{sh}}}(x).
\vspace{-0.2em}
\end{equation}
The language-specific score is obtained by averaging over the selected language-specific layers:
\vspace{-0.2em}
\begin{equation}
\label{eq:specific-score}
s_{\mathrm{sp}}^{m}(x)
=
\frac{1}{|\mathcal{L}_{\mathrm{sp}}^{m}|}
\sum_{l \in \mathcal{L}_{\mathrm{sp}}^{m}}
s_l(x).
\vspace{-0.2em}
\end{equation}

Finally, we combine the two scores to obtain the overall safety-degrading score:
\begin{equation}
\label{eq:final-score}
s(x)
=
s_{\mathrm{sh}}(x)
+
\lambda s_{\mathrm{sp}}^{m}(x),
\end{equation}
where $\lambda$ controls the contribution of language-specific sensitive layers. 
Samples exhibiting stronger unsafe affinity across sensitive layers are considered more likely to degrade safety alignment during fine-tuning.
Therefore, \textsc{MMSafe} ranks candidate fine-tuning samples by $s(x)$ and filters the highest-scoring samples as safety-degrading samples.

\section{Experiments}

\begin{table*}[t]
\centering
\footnotesize
\setlength{\tabcolsep}{1.5pt}
\renewcommand{\arraystretch}{1.06}

\begin{tabular}{l
cc cc cc cc cc cc | c}
\toprule
\textbf{Method}
& \multicolumn{6}{c}{\textbf{Aya Dataset}}
& \multicolumn{6}{c}{\textbf{Nemotron-SG}}
& \textbf{Avg.} \\
\cmidrule(lr){2-7}\cmidrule(lr){8-13}\cmidrule(lr){14-14}
& \multicolumn{2}{c}{LinguaSafe}
& \multicolumn{2}{c}{XSafety}
& \multicolumn{2}{c}{MultiJail}
& \multicolumn{2}{c}{LinguaSafe}
& \multicolumn{2}{c}{XSafety}
& \multicolumn{2}{c}{MultiJail}
&  \\
\cmidrule(lr){2-3}\cmidrule(lr){4-5}\cmidrule(lr){6-7}
\cmidrule(lr){8-9}\cmidrule(lr){10-11}\cmidrule(lr){12-13}
& \#Harm$\downarrow$ & H/R$\downarrow$
& \#Harm$\downarrow$ & H/R$\downarrow$
& \#Harm$\downarrow$ & H/R$\downarrow$
& \#Harm$\downarrow$ & H/R$\downarrow$
& \#Harm$\downarrow$ & H/R$\downarrow$
& \#Harm$\downarrow$ & H/R$\downarrow$
& H/R \\
\midrule

\multicolumn{14}{@{}l}{\textbf{LLaMA3-8B-Instruct}} \\
Random
& 4,901 & 100.0\%
& 1,810 & 100.0\%
& 719   & 100.0\%
& 2,148 & 100.0\%
& 548   & 100.0\%
& 794   & 100.0\%
& 100.0\% \\
SEAL
& 2,786 & 56.8\%
& 1,192 & 65.9\%
& 240   & 33.4\%
& 4,134 & 192.5\%
& 1,363 & 248.7\%
& 324   & 40.8\%
& 106.4\% \\
Bi-Anchor
& 3,179 & 64.9\%
& 1,378 & 76.1\%
& 521   & 72.5\%
& 3,185 & 148.3\%
& 391   & 71.4\%
& 439   & 55.3\%
& 81.4\% \\
LARF
& 3,306 & 67.5\%
& 1,337 & 73.9\%
& 463   & 64.4\%
& 571   & 26.6\%
& 36    & 6.6\%
& 79    & 9.9\%
& \underline{41.5\%} \\
\textsc{MMSafe}
& 3,348 & 68.3\%
& 1,062 & 58.7\%
& 471   & 65.5\%
& 431   & 20.1\%
& 19    & 3.5\%
& 83    & 10.5\%
& \textbf{37.8\%} \\

\midrule
\multicolumn{14}{@{}l}{\textbf{LLaMA3.1-8B-Instruct}} \\
Random
& 3,484 & 100.0\%
& 1,536 & 100.0\%
& 394   & 100.0\%
& 2,192 & 100.0\%
& 875   & 100.0\%
& 335   & 100.0\%
& 100.0\% \\
SEAL
& 4,227 & 121.3\%
& 1,347 & 87.7\%
& 288   & 73.1\%
& 2,501 & 114.1\%
& 1,206 & 137.8\%
& 268   & 80.0\%
& 102.3\% \\
Bi-Anchor
& 2,103 & 60.4\%
& 121   & 7.9\%
& 191   & 48.5\%
& 1,442 & 65.8\%
& 445   & 50.9\%
& 121   & 36.1\%
& 44.9\% \\
LARF
& 1,730 & 49.7\%
& 935   & 60.9\%
& 177   & 44.9\%
& 293   & 13.4\%
& 20    & 2.3\%
& 60    & 17.9\%
& \textbf{31.5\%} \\
\textsc{MMSafe}
& 1,824 & 52.4\%
& 937   & 61.0\%
& 185   & 47.0\%
& 155   & 7.1\%
& 20    & 2.3\%
& 70    & 20.9\%
& \underline{31.8\%} \\

\midrule
\multicolumn{14}{@{}l}{\textbf{Qwen3-8B}} \\
Random
& 3,874 & 100.0\%
& 1,581 & 100.0\%
& 419   & 100.0\%
& 3,273 & 100.0\%
& 2,071 & 100.0\%
& 308   & 100.0\%
& 100.0\% \\
SEAL
& 3,421 & 88.3\%
& 1,232 & 77.9\%
& 350   & 83.5\%
& 3,218 & 98.3\%
& 1,287 & 62.1\%
& 321   & 104.2\%
& 85.7\% \\
Bi-Anchor
& 3,065 & 79.1\%
& 1,146 & 72.5\%
& 363   & 86.6\%
& 2,094 & 64.0\%
& 432   & 20.9\%
& 222   & 72.1\%
& 65.9\% \\
LARF
& 3,196 & 82.5\%
& 1,168 & 73.9\%
& 357   & 85.2\%
& 842   & 25.7\%
& 237   & 11.4\%
& 185   & 60.1\%
& \underline{56.4\%} \\
\textsc{MMSafe}
& 3,250 & 83.9\%
& 1,014 & 64.1\%
& 312   & 74.5\%
& 742   & 22.7\%
& 76    & 3.7\%
& 92    & 29.9\%
& \textbf{46.5\%} \\

\midrule
\multicolumn{14}{@{}l}{\textbf{Qwen3-32B}} \\
Random
& 3,930 & 100.0\%
& 936   & 100.0\%
& 279   & 100.0\%
& 2,472 & 100.0\%
& 778   & 100.0\%
& 191   & 100.0\%
& 100.0\% \\
SEAL
& 3,132 & 79.7\%
& 819   & 87.5\%
& 212   & 76.0\%
& 1,978 & 80.0\%
& 642   & 82.5\%
& 111   & 58.1\%
& 77.3\% \\
Bi-Anchor
& 3,641 & 92.6\%
& 455   & 48.6\%
& 181   & 64.9\%
& 1,925 & 77.9\%
& 754   & 96.9\%
& 131   & 68.6\%
& 74.9\% \\
LARF
& 2,583 & 65.7\%
& 519   & 55.4\%
& 191   & 68.5\%
& 798   & 32.3\%
& 232   & 29.8\%
& 114   & 59.7\%
& \underline{51.9\%} \\
\textsc{MMSafe}
& 1,922 & 48.9\%
& 493   & 52.7\%
& 142   & 50.9\%
& 636   & 25.7\%
& 101   & 13.0\%
& 106   & 55.5\%
& \textbf{41.1\%} \\

\bottomrule
\end{tabular}
\caption{
Evaluation results on Aya Dataset and Nemotron-SG.
\#Harm denotes the number of harmful or safety-violating responses.
H/R denotes the harmful-response ratio relative to Random under the
same model, fine-tuning dataset, and safety benchmark:
$H/R = \#\mathrm{Harm}_{\mathrm{method}} /
\#\mathrm{Harm}_{\mathrm{Random}} \times 100\%$.
Avg. reports the average H/R across the six benchmark--dataset combinations.
The best Avg. result within each model group is shown in \textbf{bold}, and the second-best is \underline{underlined}.
}
\vspace{-1.2em}
\label{tab:safety_eval}
\end{table*}

\begin{table*}[t]
\centering
\small
\begin{tabular}{llrrrr}
\toprule
\textbf{Model} & \textbf{FT Data} &
\textbf{MATH} & \textbf{Code} &
\textbf{XNLI} & \textbf{MLQA} \\
\midrule
Qwen3-8B
& Base        & 90.4 & 52.5 & 71.09 & 61.41 \\
& Aya Dataset & 90.5 & 51.5 & 74.51 & 64.83 \\
& Nemotron-SG & 89.6 & 52.0 & 72.32 & 62.32 \\
\midrule
Qwen3-32B
& Base        & 92.2 & 60.7 & 70.95 & 62.94 \\
& Aya Dataset & 91.7 & 59.9 & 72.56 & 65.25 \\
& Nemotron-SG & 91.1 & 60.1 & 72.15 & 63.86 \\
\midrule
LLaMA3-8B-Instruct
& Base        & 27.1 & 55.4 & 42.52 & 58.79 \\
& Aya Dataset & 26.2 & 55.2 & 45.31 & 60.23 \\
& Nemotron-SG & 27.3 & 54.7 & 43.14 & 59.27 \\
\midrule
LLaMA3.1-8B-Instruct
& Base        & 45.9 & 67.6 & 49.74 & 58.28 \\
& Aya Dataset & 46.2 & 67.2 & 51.58 & 59.47 \\
& Nemotron-SG & 45.2 & 67.4 & 49.83 & 57.99 \\
\bottomrule
\end{tabular}
\caption{Utility evaluation after fine-tuning with data selected by
MMSAFE. MATH and Code evaluate general reasoning and programming
capabilities, while XNLI and MLQA evaluate multilingual natural
language understanding and question answering. For Code, Qwen models
are evaluated on LiveCodeBench and LLaMA models on HumanEval.}
\label{tab:utility-results}
\vspace{-2em}
\end{table*}

\begin{figure*}[t]
  \centering
  \begin{minipage}[t]{0.62\textwidth}
    \centering
    \includegraphics[width=\linewidth]{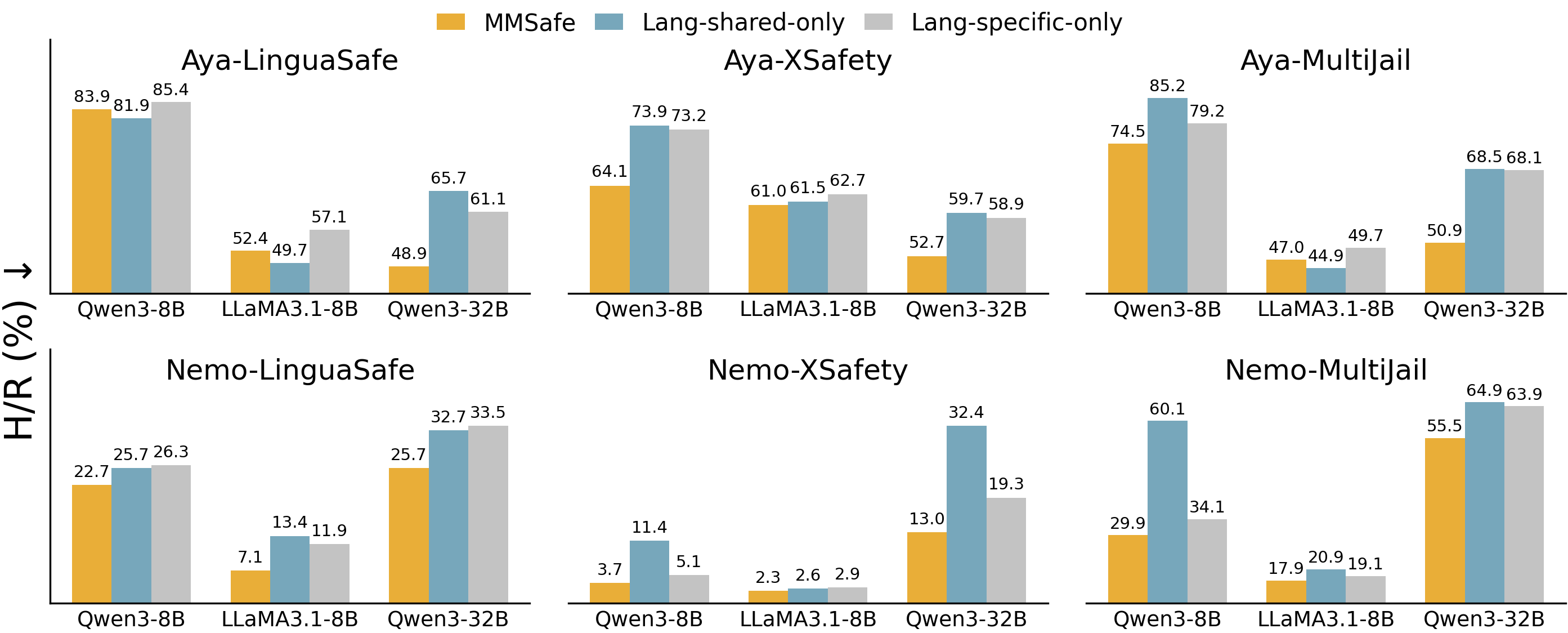}
    \captionof{figure}{%
    Ablation study on language-shared and language-specific sensitive layers.
    Lang-shared-only uses only the cross-lingually shared sensitive layer, Lang-specific-only uses only language-specific sensitive layers, and \textsc{MMSafe} combines both.
    We report H/R relative to Random across three models and six benchmark--dataset settings.
    Lower values indicate fewer harmful responses.
    }
    \vspace{-1.2em}
    \label{fig:ablation_shared_specific}
  \end{minipage}\hfill
  \begin{minipage}[t]{0.36\textwidth}
    \centering
    \includegraphics[width=\linewidth]{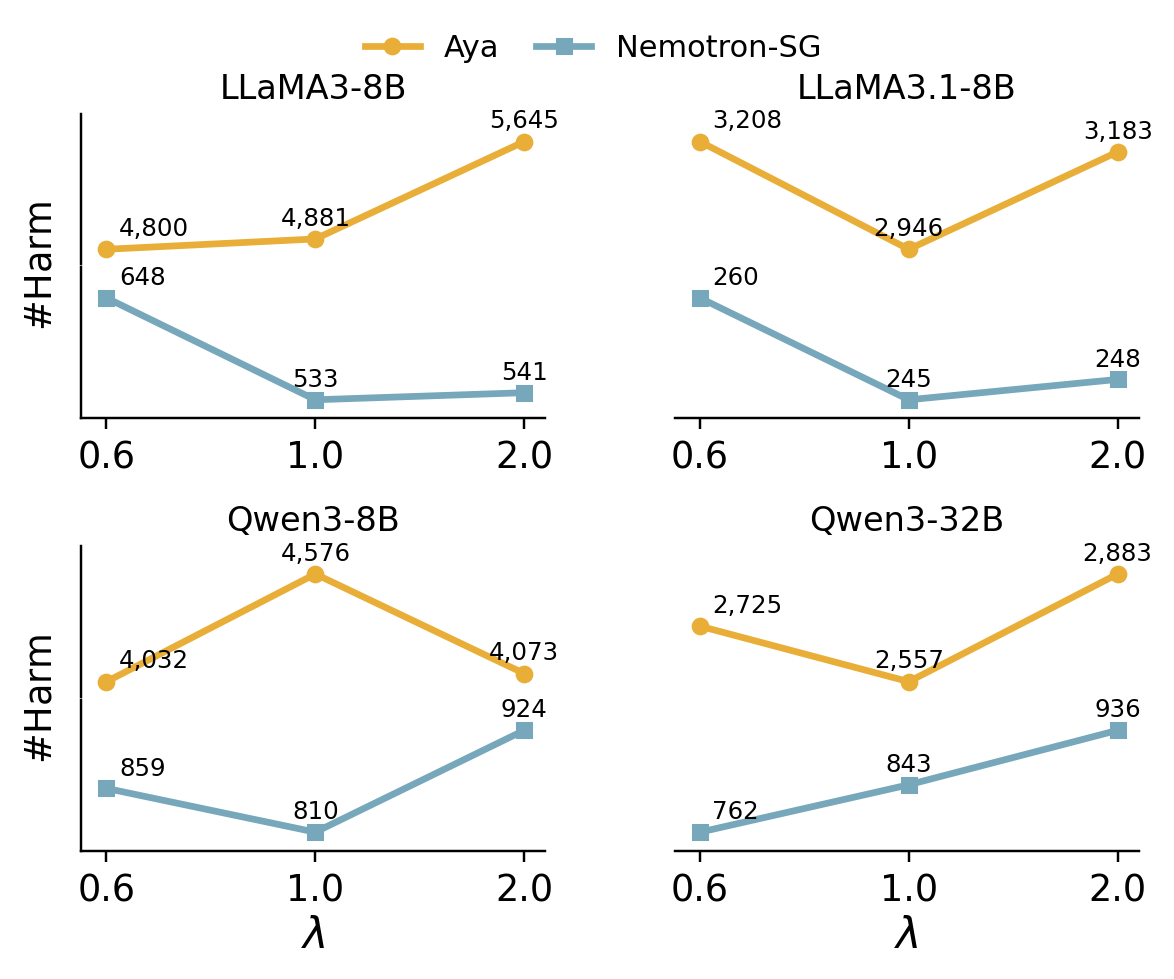}
    \captionof{figure}{%
    Sensitivity analysis of the weighting coefficient $\lambda$ in \textsc{MMSafe}.
    For each model, we report the summed number of harmful responses on Aya Dataset and Nemotron-SG across three safety benchmarks.
    }
    \vspace{-1.2em}
    \label{fig:lambda_sensitivity}
  \end{minipage}
\end{figure*}

\subsection{Experimental Setup}

\paragraph{Models.}
We evaluate \textsc{MMSafe} on four large language models: LLaMA3-8B-Instruct \citep{DBLP:journals/corr/abs-2407-21783}, LLaMA3.1-8B-Instruct \citep{DBLP:journals/corr/abs-2407-21783}, Qwen3-8B \citep{DBLP:journals/corr/abs-2505-09388}, and Qwen3-32B \citep{DBLP:journals/corr/abs-2505-09388}.
These models cover different model families and scales, allowing us to examine whether \textsc{MMSafe} consistently identifies safety-degrading samples across diverse model architectures.

\paragraph{Datasets and Benchmarks.}
We use Aya Dataset \citep{DBLP:conf/acl/SinghVD0MKSPMOZ24} and the safe-labeled subset of Nemotron-Safety-Guard-v3 \citep{DBLP:conf/ijcnlp/JoshiPSKELGVLCW25} as candidate fine-tuning data pools, totaling over 400K samples and covering up to 65 languages.
For safety evaluation, we use three multilingual safety benchmarks: LinguaSafe \citep{DBLP:journals/corr/abs-2508-12733}, XSafety \citep{wang-etal-2024-languages}, and MultiJail \citep{DBLP:conf/iclr/0010ZPB24}, covering over 70K evaluation instances across 16 languages.
Additional datasets used for sensitive layer discovery and safe-unsafe centroid construction are described in Appendix~\ref{app:experiment-setting}.

\paragraph{Baselines.}
We compare \textsc{MMSafe} with four baselines.
\emph{Random} uniformly samples data from the candidate pool.
\emph{Bi-Anchoring} \citep{heyour2024} estimates sample risk by measuring gradient similarity between candidate samples and reference instances.
\emph{SEAL} \citep{DBLP:conf/iclr/ShenCDC25} trains a ranking model to distinguish safe and unsafe samples.
\emph{LARF} \citep{DBLP:conf/emnlp/LiLLWLSS25} identifies safety-degrading samples using representations from a single safety-sensitive layer.

\paragraph{Implementation Details.}
We set the weighting coefficient in Equation~\ref{eq:final-score} to $\lambda=1$.
For each filtering method, we fine-tune models on the bottom 20\% lowest-risk samples selected by each method using LoRA.
For all generation tasks, we use deterministic decoding with temperature $0$ to ensure reproducibility.
Additional implementation details are provided in Appendix~\ref{app:experiment-setting}.

\paragraph{Evaluation.}
We primarily use LlamaGuard~3 to evaluate model safety after fine-tuning, following prior safety-degrading data filtering work for consistent comparison with LARF. We further validate the robustness of our evaluation with additional automatic and human judges, which yield similar safety trends; see Appendix~\ref{app:evaluator_robustness}.

\subsection{Main Results}

Table~\ref{tab:safety_eval} shows the main safety evaluation results. 
We report both the number of harmful responses (\#Harm) and the harmful-response ratio (H/R) relative to Random selection under the same setting. 
Lower values indicate fewer safety-violating responses after fine-tuning.

\textsc{MMSafe} consistently achieves the strongest or near-strongest safety performance across model families.
In particular, \textsc{MMSafe} outperforms LARF, the single-layer baseline, on three out of four models, reducing the average H/R from 41.5\% to 37.8\%, from 56.4\% to 46.5\%, and from 51.9\% to 41.1\% on LLaMA3-8B-Instruct, Qwen3-8B, and Qwen3-32B, respectively.
These improvements suggest that combining language-shared and language-specific sensitive layers can provide a more effective signal for identifying safety-degrading samples than relying on a single layer.

Across fine-tuning datasets and safety benchmarks, \textsc{MMSafe} consistently reduces harmful responses compared with Random selection, indicating that it can filter samples that would otherwise induce safety degradation during fine-tuning. 
These results highlight the importance of modeling multilingual safety signals across multiple layers rather than relying on a single-layer view.

\subsection{Utility Preservation}

We further evaluate whether \textsc{MMSafe} preserves both general and multilingual capabilities after safety-aware filtering.
For general utility, we evaluate mathematical reasoning on MATH \citep{lightman2023let} and code generation using LiveCodeBench \citep{DBLP:conf/iclr/JainHGLYZWSSS25} for Qwen3 models and HumanEval \citep{DBLP:journals/corr/abs-2107-03374} for LLaMA models.
To complement these benchmarks, we additionally evaluate multilingual utility on XNLI \citep{conneau-etal-2018-xnli} for cross-lingual natural language inference and MLQA \citep{lewis-etal-2020-mlqa} for multilingual question answering.

As shown in Table~\ref{tab:utility-results}, \textsc{MMSafe}-selected
fine-tuning largely preserves general utility while maintaining or
improving multilingual performance.
MATH and code-generation scores remain close to those of the base
models across both Aya Dataset and Nemotron-SG.
For multilingual utility, fine-tuning on Aya improves all four models
on both XNLI and MLQA, indicating that the selected data can preserve
or even enhance cross-lingual understanding.
Nemotron-SG also preserves or slightly improves performance in most
settings, with only a small decrease on MLQA for
LLaMA3.1-8B-Instruct (58.28 to 57.99).
The improvements are particularly consistent on Aya, where all models
show gains across both multilingual benchmarks.
This suggests that the benefits of \textsc{MMSafe} are not limited to
safety preservation, but can coexist with stable downstream capability
across different fine-tuning data distributions.
Overall, these results suggest that the safety improvements achieved
by \textsc{MMSafe} do not come at the cost of substantial degradation
in either general or multilingual utility.

\section{Analysis}
\subsection{Ablation on Shared and Language-Specific Layers}

Figure~\ref{fig:ablation_shared_specific} compares \textsc{MMSafe} with two ablated variants that use only the language-shared sensitive layer or only language-specific sensitive layers. The results show that neither component alone consistently dominates across models and datasets. The shared-only variant performs competitively in several LLaMA3.1-8B settings, indicating that shared layers encode transferable multilingual safety signals. However, it becomes less stable on Qwen models, especially under Nemotron-SG evaluations. In contrast, the language-specific-only variant improves some multilingual settings, but still underperforms the full method in several cases.

These results indicate that the two types of layers provide complementary information. Shared layers offer a stable cross-lingual anchor, while language-specific layers capture safety patterns that vary across languages. By combining both, \textsc{MMSafe} achieves more robust safety filtering across heterogeneous multilingual settings.

\subsection{Sensitivity to Language-Specific Weight}

Figure~\ref{fig:lambda_sensitivity} analyzes the effect of the weighting coefficient $\lambda$, which controls the contribution of language-specific layers. The optimal value varies across models and datasets. For example, on LLaMA3-8B, increasing $\lambda$ reduces harmful responses on Nemotron-SG but increases them on Aya. Qwen3-32B shows a similar trade-off, with different datasets exhibiting different preferences for language-specific weighting. This suggests that over-emphasizing language-specific layers may benefit some data distributions but hurt others.

Overall, $\lambda=1.0$ provides a stable middle ground. Although it is not always optimal for every individual model--dataset pair, it avoids extreme behavior and balances shared and language-specific signals. Therefore, we use $\lambda=1$ as the default setting in the main experiments.

\begin{figure}[t]
  \centering
  \includegraphics[width=\linewidth]{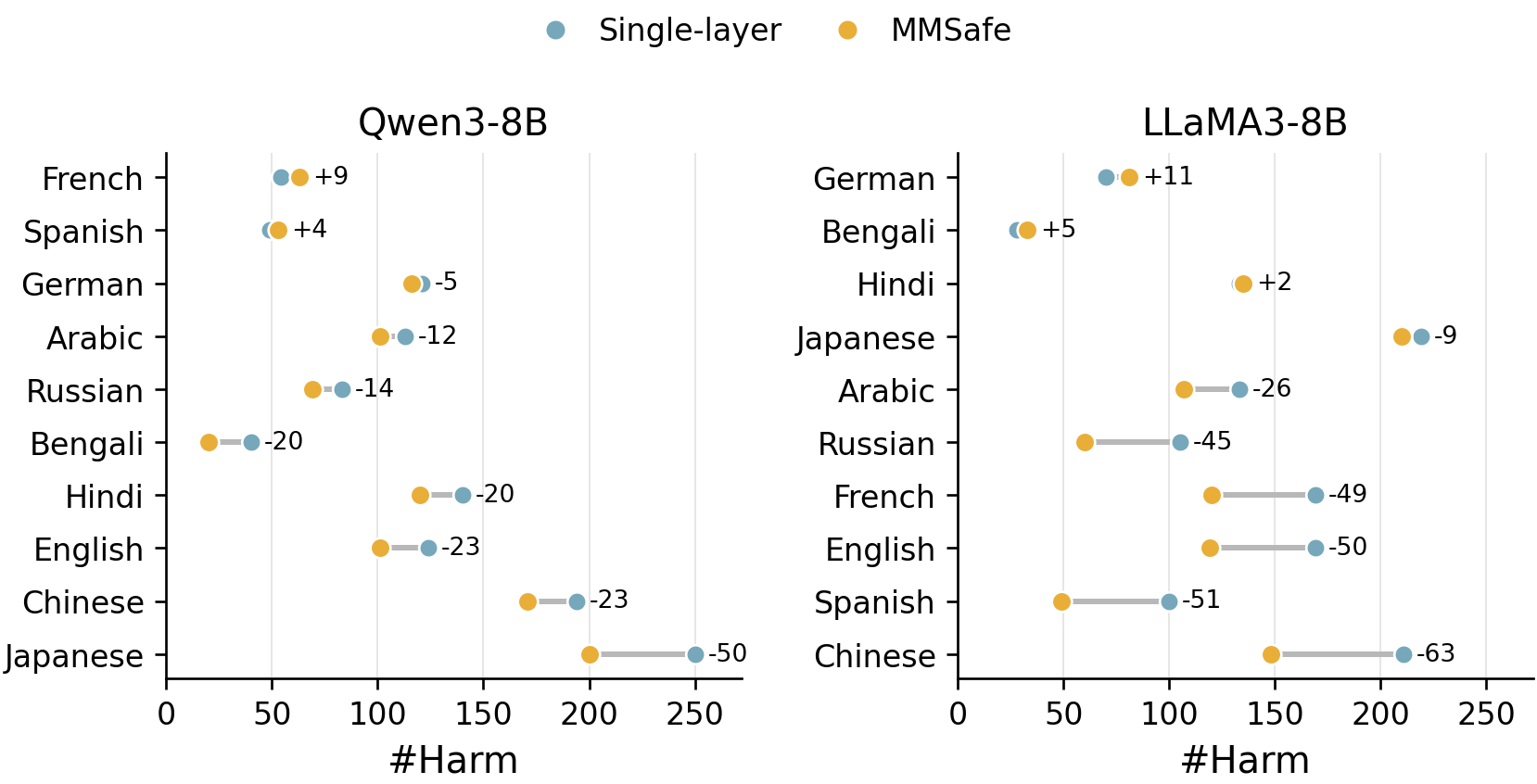}
  \vspace{-1.2em}
  \caption{
  Language-wise unsafe counts on XSafety. 
  Negative changes indicate fewer harmful responses with \textsc{MMSafe}.
  }
  \vspace{-1.2em}
  \label{fig:language_dumbbell_xsafety}
\end{figure}

\begin{figure}[t]
  \centering
  \includegraphics[width=\linewidth]{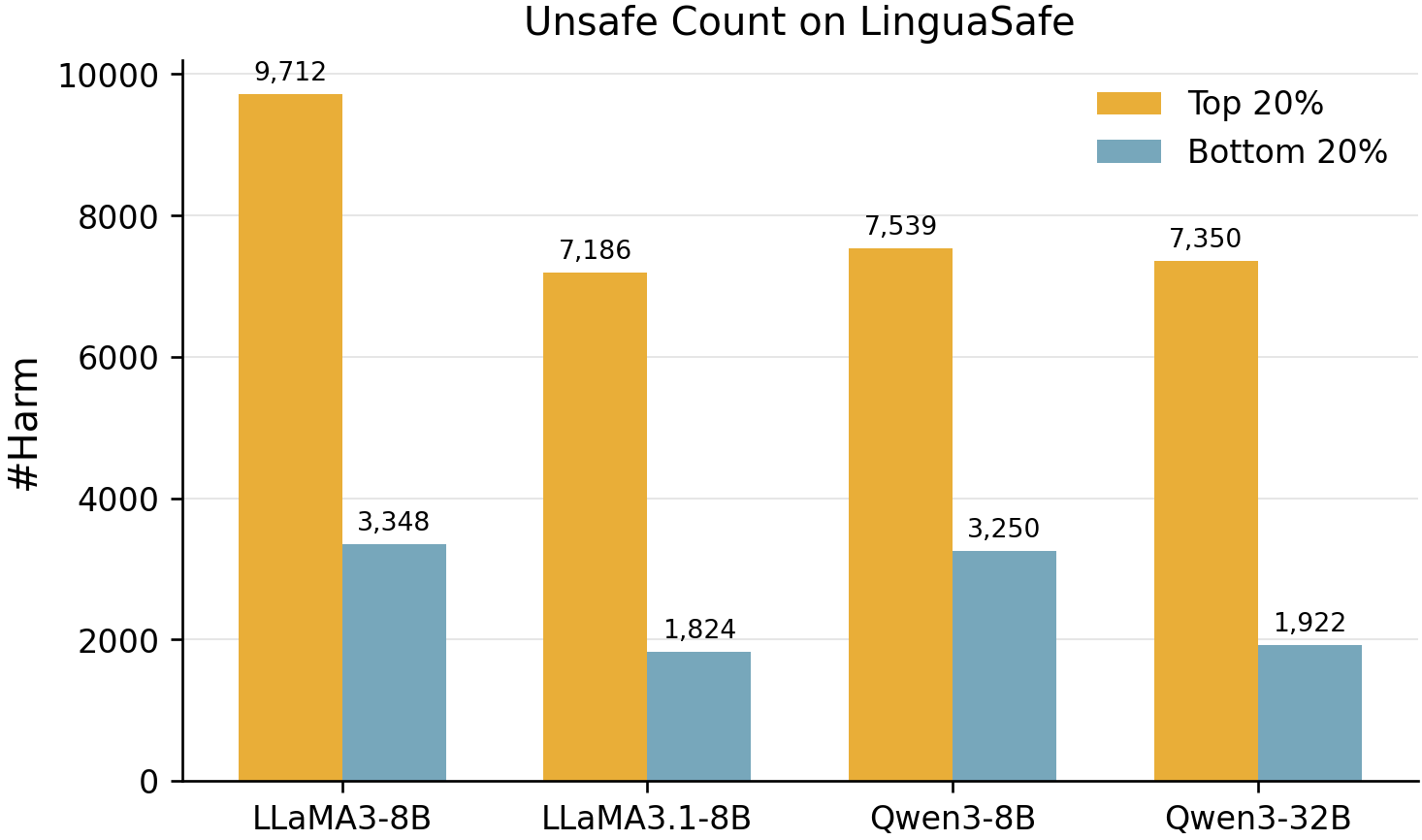}
  \vspace{-1.2em}
  \caption{
  Unsafe counts on LinguaSafe after fine-tuning with top-20\% vs. bottom-20\% ranked Aya dataset.
  }
  \vspace{-1.2em}
  \label{fig:linguasafe_top_bottom}
\end{figure}

\subsection{Language-wise Safety Analysis}

Figure~\ref{fig:language_dumbbell_xsafety} shows language-wise unsafe counts on Aya-XSafety. 
Overall, \textsc{MMSafe} reduces harmful responses in most languages for both Qwen3-8B and LLaMA3-8B, showing that the improvement is not driven by only a few high-resource languages. 
The reduction is especially clear for languages such as Japanese, Chinese, English, and Russian. 
Although a few languages show small regressions, the overall trend demonstrates MMSAFE’s consistent robustness across diverse language groups.

\subsection{Effect of Ranked Sample Selection}

Figure~\ref{fig:linguasafe_top_bottom} compares fine-tuning on the top-20\% and bottom-20\% ranked Aya samples.
Across all four models, the top-ranked samples lead to substantially more unsafe responses on LinguaSafe than the bottom-ranked samples.
This supports that our scoring function assigns higher ranks to samples with greater potential to degrade safety during fine-tuning, and that the improvement from bottom-20\% selection is not merely due to the filtering ratio itself.

We further inspect high-ranked samples and find that they often contain open-ended or fictionalized negative scenarios involving disasters, crime, or violence, as well as benign-looking content with safety-sensitive lexical cues.
These patterns are consistent with the safety-relevant features identified by our SAE analysis, including lexical harmful cues, concrete harmful scenarios, harmful intent or deception, and broader safety-policy violations.
Importantly, such samples are not necessarily explicitly unsafe, suggesting that MMSAFE captures safety-degrading potential beyond surface-level harmfulness.

\section{Related Work}
\subsection{Safety Degradation under Fine-tuning}

Recent studies show that safety alignment in LLMs can be fragile during downstream fine-tuning.
Even benign-looking data may weaken safety safeguards and increase harmful responses after adaptation \citep{Qi0XC0M024,heyour2024,DBLP:conf/icml/Halawi0WWHS24}.
This has motivated optimization-based defenses \citep{DBLP:conf/nips/HuangHIT024} and data-centric selection or filtering methods \citep{DBLP:conf/iclr/ShenCDC25,DBLP:conf/emnlp/LiLLWLSS25}.
However, most prior methods focus on monolingual or language-agnostic settings, and representation-based filtering typically relies on a single safety-sensitive layer.
Our work instead studies multilingual safety-degrading data identification and shows that safety-relevant signals are distributed across multiple layers.

\subsection{Multilingual LLM Safety}

Multilingual safety has recently received increasing attention, as LLMs may exhibit uneven safety behavior across languages.
Prior work has revealed multilingual jailbreak vulnerabilities and stronger safety risks for non-English prompts \citep{DBLP:conf/iclr/0010ZPB24,wang-etal-2024-languages}.
Recent benchmarks, datasets, and guard models further support multilingual safety evaluation and moderation \citep{DBLP:conf/iclr/0010ZPB24,wang-etal-2024-languages,kumar2025polyguard,DBLP:conf/ijcnlp/JoshiPSKELGVLCW25,DBLP:conf/icml/MazeikaPYZ0MSLB24,DBLP:journals/corr/abs-2508-12733}.
In contrast to prior work that mainly evaluates multilingual safety after model deployment, our work focuses on multilingual safety-degrading data identification before fine-tuning.

\section{Conclusion}

In this paper, we study safety-degrading data identification in multilingual fine-tuning. 
We show that safety-sensitive layers are only partially shared across languages, and that safety-relevant signals are distributed across both shared and language-specific layers. 
Motivated by these findings, we propose \textsc{MMSafe}, a multilingual multi-layer safety filter that jointly leverages language-shared and language-specific sensitive layers to score candidate fine-tuning samples.
Experiments across multiple models, data pools, and multilingual safety benchmarks show that \textsc{MMSafe} consistently reduces harmful responses after fine-tuning while largely preserving general utility.

\section*{Limitations}

MMSAFE is a pre-fine-tuning filtering method rather than a complete
safety solution, and adaptive or filtering-aware attacks may still
circumvent the selected layers or scoring rule. In addition, sensitive-layer
discovery relies on language-specific refusal patterns, while the
safe--unsafe prototypes depend on the coverage of the reference data;
both may introduce language-dependent bias, especially for lower-resource
or unseen languages. Although we validate the main trends with multiple
automatic and human evaluators, automatic and translation-based safety
judgments may still exhibit language- or culture-dependent bias.

\bibliography{custom}

\appendix

\section{Experiment Setting}
\label{app:experiment-setting}
\subsection{Dataset Usage}

\begin{table*}[t]
\centering
\small
\setlength{\tabcolsep}{5pt}
\renewcommand{\arraystretch}{1.12}
\begin{tabular}{llll}
\toprule
Category & Dataset & Languages & Size / Coverage \\
\midrule
Candidate fine-tuning data 
& Aya Dataset 
& 65 
& 204K samples \\

Candidate fine-tuning data 
& Nemotron-SG 
& 12 
& $>$200K safe-labeled samples \\

Sensitive layer discovery 
& OR-Bench 
& 18$^\dagger$ 
& 1K hard queries translated into multiple languages \\

Sensitive layer discovery 
& PolyGuardMix 
& 17 
& 1.91M samples \\

Safe-unsafe centroid construction 
& PolyGuardMix 
& 17 
& 1.91M samples \\

Safe-unsafe centroid construction 
& Nemotron-SG Val. 
& 12 
& $\sim$10K validation samples \\

Safety evaluation 
& LinguaSafe 
& 12 
& 45K entries \\

Safety evaluation 
& XSafety 
& 10 
& 28K samples covering 14 safety categories \\

Safety evaluation 
& MultiJail 
& 10 
& Multilingual jailbreak prompts \\
\bottomrule
\end{tabular}
\caption{
Summary of datasets and benchmarks used in our experiments.
Nemotron-SG denotes Nemotron-Safety-Guard-v3.
$^\dagger$ indicates languages obtained through translation.
}
\label{tab:dataset-summary}
\end{table*}

\begin{table*}[t]
\centering
\small
\setlength{\tabcolsep}{4pt}
\renewcommand{\arraystretch}{1.12}
\begin{tabular}{l c p{0.68\textwidth}}
\toprule
Model & Shared layer & Language-specific layers \\
\midrule
Qwen3-8B
& 19
& Arabic: 27, 17; Czech: 28, 14; German: 14, 24; English: 14, 13; Spanish: 14, 17; French: 26, 14; Hindi: 17, 24; Italian: 13, 20; Japanese: 24, 21; Korean: 29, 27; Dutch: 13, 24; Thai: 22, 5; Chinese: 15, 22; Portuguese: 29, 17; Russian: 28, 29; Polish: 24, 17; Swedish: 24, 29. \\

\midrule
LLaMA3-8B-Instruct
& 9
& Arabic: 10, 0; Czech: 13, 2; German: 29, 23; English: 10, 14; Spanish: 13, 11; French: 28, 11; Hindi: 0, 2; Italian: 2, 3; Japanese: 10, 0; Korean: 10, 0; Dutch: 13, 10; Thai: 13, 10; Chinese: 10, 0; Portuguese: 13, 11; Russian: 10, 13; Polish: 2, 10; Swedish: 2, 1. \\

\midrule
LLaMA3.1-8B-Instruct
& 8
& Arabic: 10, 2; Czech: 28, 2; German: 4, 19; English: 2, 15; Spanish: 15, 10; French: 15, 10; Hindi: 20, 2; Italian: 15, 2; Japanese: 12, 14; Korean: 12, 14; Dutch: 18, 19; Thai: 10, 9; Chinese: 10, 15; Portuguese: 15, 10; Russian: 20, 2; Polish: 2, 15; Swedish: 2, 4. \\

\midrule
Qwen3-32B
& 42
& Arabic: 52, 51; Czech: 51, 55; German: 44, 37; English: 44, 40; Spanish: 52, 56; French: 44, 58; Hindi: 57, 54; Italian: 51, 52; Japanese: 20, 40; Korean: 59, 50; Dutch: 61, 45; Thai: 52, 51; Chinese: 44, 46; Portuguese: 52, 59; Serbian: 1, 56; Russian: 54, 41; Polish: 54, 56; Swedish: 56, 54. \\
\bottomrule
\end{tabular}
\caption{
Selected language-shared and language-specific sensitive layers used by \textsc{MMSafe}.
For each model, the shared layer is selected by aggregating sensitivity across languages, and the language-specific layers are selected from each language's top-ranked layers after excluding the shared layer.
}
\label{tab:selected-sensitive-layers}
\end{table*}

\begin{figure*}[t]
  \centering
  \includegraphics[width=\textwidth]{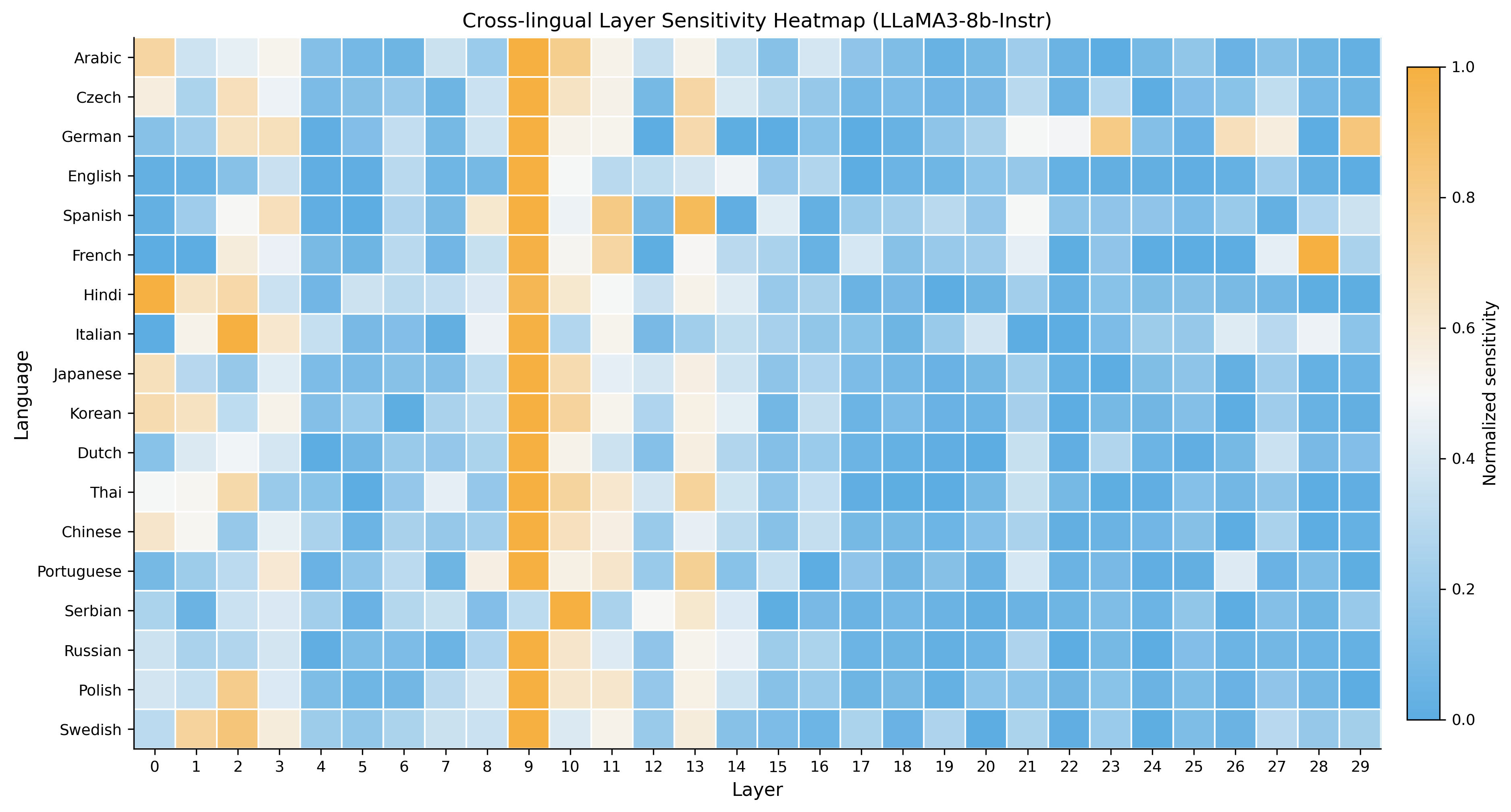}
  \vspace{-0.8em}
  \caption{
Cross-lingual layer sensitivity of LLaMA3-8B-Instruct across 18 languages.
Each row is normalized within a language, and darker orange indicates higher sensitivity.
}
  \vspace{-0.5em}
  \label{fig:LLAMA3_8B_sensitivity_heatmap}
\end{figure*}

\begin{figure*}[t]
  \centering
  \includegraphics[width=\textwidth]{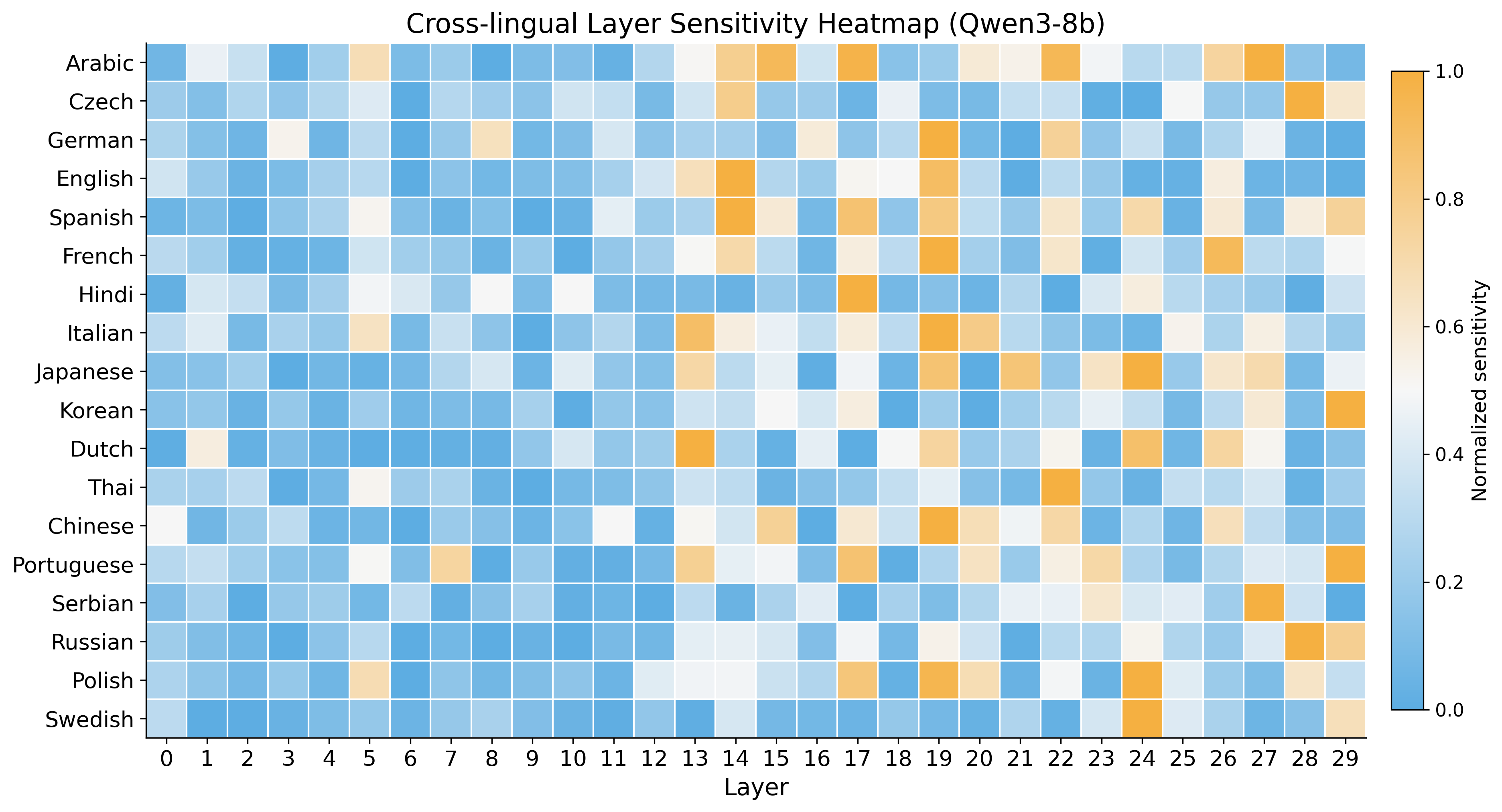}
  \vspace{-0.8em}
  \caption{
Cross-lingual layer sensitivity of Qwen3-8B across 18 languages.
Each row is normalized within a language, and darker orange indicates higher sensitivity.
}
  \vspace{-0.5em}
  \label{fig:QWEN3_8B_sensitivity_heatmap}
\end{figure*}

\begin{figure*}[t]
  \centering
  \includegraphics[width=\textwidth]{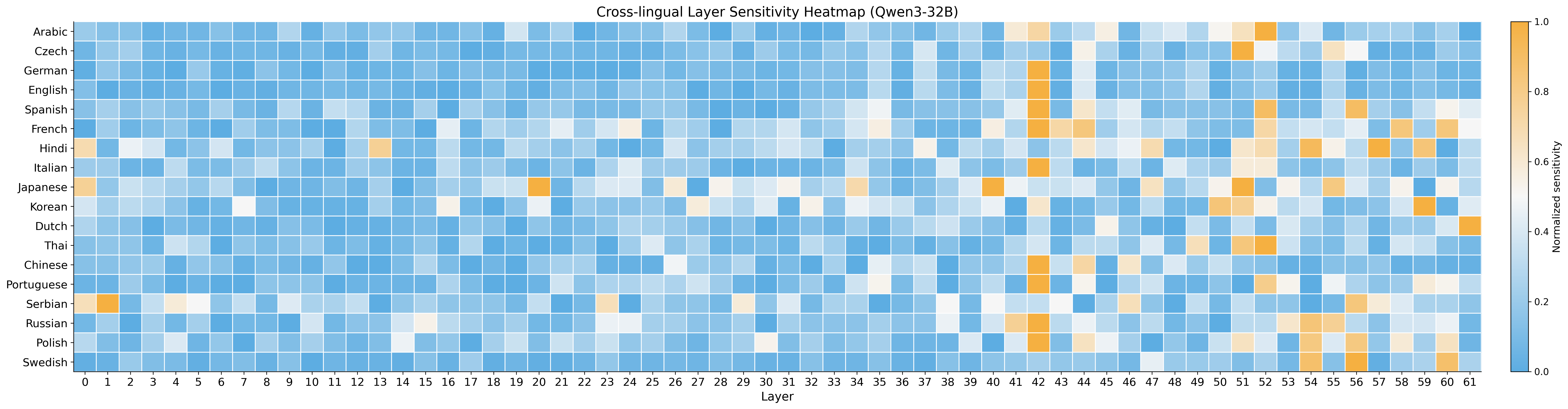}
  \vspace{-0.8em}
  \caption{
Cross-lingual layer sensitivity of Qwen3-32B across 18 languages.
Each row is normalized within a language, and darker orange indicates higher sensitivity.
}
  \vspace{-0.5em}
  \label{fig:Qwen3_32B_sensitivity_heatmap}
\end{figure*}

We use multiple datasets for different purposes in our experiments, including candidate fine-tuning data pools, multilingual safety evaluation benchmarks, sensitive layer discovery data, and reference data for safe-unsafe centroid construction.

\paragraph{Candidate fine-tuning data pools.}
We use Aya Dataset and the safe-labeled subset of Nemotron-Safety-Guard-v3 as candidate fine-tuning data pools for safety-degrading sample identification. 
Aya Dataset is a large-scale multilingual instruction-following dataset containing 204K samples across 65 languages. 
We use it as a broad multilingual fine-tuning corpus, where safety-degrading samples may be mixed with otherwise benign instruction-following data. 
We further use the safe-labeled subset of Nemotron-Safety-Guard-v3, which contains over 200K safe-labeled samples across 12 languages. 
Although these samples are labeled as safe, prior work has shown that seemingly benign fine-tuning data can still contain examples that degrade safety alignment, making this dataset a challenging testbed for safety-degrading sample identification.

\paragraph{Safety evaluation benchmarks.}
We evaluate safety preservation after fine-tuning on three multilingual safety benchmarks: LinguaSafe, XSafety, and MultiJail. 
LinguaSafe contains 45K entries across 12 languages and is designed to evaluate multilingual safety behavior. 
XSafety covers 14 commonly used safety issue categories across 10 languages, allowing us to evaluate safety under diverse harmful request types. 
MultiJail evaluates multilingual jailbreak robustness across 10 languages. 
Together, these benchmarks cover a broad range of languages, safety categories, and jailbreak scenarios.

\paragraph{Sensitive layer discovery data.}
For sensitive layer discovery, we construct a multilingual harmful query set from OR-Bench and PolyGuardMix. 
Specifically, we use 2K harmful queries for each of 18 languages, resulting in 36K queries in total. 
These queries are used to perturb each layer and compute language-wise sensitivity scores for selecting language-shared and language-specific sensitive layers.

\paragraph{Reference safe and unsafe pairs.}
To construct the safe and unsafe centroids used by \textsc{MMSafe}, we sample reference question-answer pairs from PolyGuardMix and the validation split of Nemotron-Safety-Guard-v3. 
For each of the 18 languages, we sample 800 safe and 800 unsafe pairs. 
The safe pairs are used to estimate layer-wise safe centroids, while the unsafe pairs are used to estimate layer-wise unsafe centroids for safe-unsafe contrastive scoring.

\subsection{Implementation Details}

\paragraph{Sensitive layer discovery.}
For sensitive layer discovery, we follow the perturbation-based sensitivity measurement of LARF. 
For each language, we use 2K harmful queries, covering 18 languages and 36K queries in total. 
We perturb each transformer layer with scaling factors $0.8$ and $1.2$, corresponding to a perturbation strength of $\alpha=0.2$. 
For each perturbed model, we generate responses to the harmful queries and compute the safety response count difference between the two scaling directions as the layer sensitivity score.

To reduce computational cost, we set \texttt{max\_new\_tokens} to 32 during sensitive layer discovery. 
Following prior work, we observe that the model's refusal intent can usually be inferred from the initial response tokens. 
We therefore detect refusals by matching predefined refusal-related patterns, such as ``I cannot'', ``I am sorry'', and ``Sorry'', in the generated responses. For non-English responses, we include language-specific refusal patterns translated and manually checked for each evaluated language.

\paragraph{Safe-unsafe centroid construction.}
For constructing the safe and unsafe centroids used by \textsc{MMSafe}, we use reference question-answer pairs from PolyGuardMix and the validation split of Nemotron-Safety-Guard-v3. 
For each of the 18 languages, we sample 800 safe and 800 unsafe question-answer pairs. 
Given the selected sensitive layers, we extract layer-wise representations for these reference pairs and compute the mean representations as safe and unsafe centroids. 
These centroids are then used for safe-unsafe contrastive scoring with cosine similarity.

\paragraph{Layer selection and scoring.}
For each model, we select one language-shared sensitive layer by aggregating sensitivity scores across languages. 
For each language, we select the top-$k$ language-specific sensitive layers after excluding the language-shared layer, and set $k=2$ in all experiments. 
We set the weighting coefficient in Equation~\ref{eq:final-score} to $\lambda=1$. 
The final safety-degrading score is computed by combining the shared-layer score and the language-specific score.

\paragraph{Filtering and fine-tuning.}
For each candidate fine-tuning data pool and each filtering method, we rank candidate samples according to their safety-degrading scores and select the bottom 20\% samples for fine-tuning. 
We use the same selection ratio for all methods to ensure a fair comparison. 
We adopt LoRA fine-tuning for all experiments, with \texttt{num\_train\_epochs} set to 3, learning rate set to $1\times10^{-4}$, and warmup ratio set to 0.1. 
Using 20\% of the original data provides a controlled filtering setting: it retains sufficient fine-tuning signal while requiring each method to prioritize samples that are most likely to affect safety alignment.

\paragraph{Generation.}
For all generation tasks, we set \texttt{do\_sample} to
\texttt{False} and temperature to 0 to ensure reproducibility.
Unless otherwise specified, the same generation settings are used
across models, datasets, and filtering methods.

\paragraph{Computational cost.}
Most of the additional cost of \textsc{MMSafe} comes from one-time
offline preparation, including sensitive-layer discovery and
safe--unsafe prototype construction for each base model.
Once constructed, these components can be reused across candidate
datasets, and subsequent filtering only requires representation
extraction from the selected layers and similarity-based scoring.
All experiments were conducted on NVIDIA A100 GPUs.
Sensitive-layer discovery, LoRA fine-tuning, and evaluation required
approximately 100 GPU hours in total.

\subsection{Robustness of Safety Evaluation}
\label{app:evaluator_robustness}

Our main experiments use LlamaGuard~3 as the safety evaluator, following the evaluation protocol adopted by prior work on safety-degrading data filtering and enabling a consistent comparison with LARF. 
However, multilingual safety evaluation can be sensitive to the choice of automatic judge. 
To examine whether our conclusions depend on LlamaGuard~3, we conduct an additional validation study using three complementary evaluation protocols: GPT-4 judgment, translate-then-LlamaGuard, and human judgment.

\paragraph{Validation setup.}
For each fine-tuned model, we randomly sample 1,200 model responses from the main safety evaluation results. 
Specifically, we sample 200 responses from each combination of the three safety benchmarks---LinguaSafe, XSafety, and MultiJail---and the two fine-tuning datasets---Aya Dataset and Nemotron-SG. 
This yields 4,800 responses in total across the four evaluated models.

We evaluate the sampled responses using four protocols.
First, \textbf{LlamaGuard~3} directly classifies the original multilingual responses as safe or unsafe, following the protocol used in our main experiments.
Second, \textbf{GPT-4} directly judges whether each response contains harmful or safety-violating content.
Third, for \textbf{translate-then-LlamaGuard}, we first translate non-English responses into English using Gemini-3 and then apply LlamaGuard~3 to the translated responses.
Finally, for \textbf{human judgment}, four annotators each label 1,200 translated responses as safe or unsafe, covering the full set of 4,800 sampled responses.

\begin{table}[t]
\centering
\small
\begin{tabular*}{\columnwidth}{@{\extracolsep{\fill}}lrrrr@{}}
\toprule
\textbf{Model} &
\textbf{LG-3} &
\textbf{GPT-4} &
\textbf{T-LG} &
\textbf{Human} \\
\midrule
Qwen3-8B    & 532 & 518 & 535 & 528 \\
Qwen3-32B   & 504 & 496 & 509 & 481 \\
LLaMA3-8B   & 445 & 430 & 441 & 432 \\
LLaMA3.1-8B & 391 & 387 & 388 & 385 \\
\bottomrule
\end{tabular*}

\caption{Unsafe response counts under different safety evaluation
protocols. Each model is evaluated on 1,200 sampled responses.
\textbf{LG-3} denotes LlamaGuard~3, and \textbf{T-LG} denotes
translate-then-LlamaGuard.}
\label{tab:evaluator_robustness}
\vspace{-1.2em}
\end{table}

\paragraph{Results.}
As shown in Table~\ref{tab:evaluator_robustness}, the aggregate unsafe-response counts are similar across the four evaluation protocols.
More importantly, all evaluators produce the same model-level ordering: Qwen3-8B exhibits the largest number of unsafe responses, followed by Qwen3-32B, LLaMA3-8B, and LLaMA3.1-8B.
For example, LlamaGuard~3 identifies 532 unsafe responses for Qwen3-8B, compared with 518 by GPT-4, 535 by translate-then-LlamaGuard, and 528 by human judgment.
A similar pattern holds for the other three models.

These results suggest that the safety trends observed in our main experiments are not specific to the use of LlamaGuard~3 as the evaluator.
Nevertheless, automatic and translation-based safety evaluation may still introduce language- or culture-dependent biases, and we therefore regard multi-evaluator validation as complementary rather than a complete substitute for broader native-language human evaluation.

\section{Sensitive Layers Further Analysis}
\label{app:sensitive-further-analysis}
\subsection{Multilingual Harmful Query Set}

To analyze layer sensitivity across languages, we construct a multilingual harmful query set from two sources: OR-Bench and PolyGuardMix. 
From OR-Bench, we select 1,000 hard safety-related queries and translate them into multiple languages using Gemini. 
To complement these translated queries with naturally multilingual harmful data, we further incorporate samples from PolyGuardMix.

For each language, the resulting evaluation set contains 2,000 harmful queries, consisting of translated OR-Bench queries and native multilingual queries from PolyGuardMix. 
In total, our sensitivity analysis covers 18 languages and 36K harmful queries. 
The translated OR-Bench queries provide controlled cross-lingual counterparts, while PolyGuardMix contributes naturally occurring multilingual harmful queries.
This construction allows us to evaluate both cross-lingual consistency and language-specific variation in layer sensitivity.

\subsection{Sensitivity Patterns Across Additional Models}

We further examine whether the partially shared sensitivity pattern observed in the main text generalizes to other models. 
Figures~\ref{fig:LLAMA3_8B_sensitivity_heatmap}, \ref{fig:QWEN3_8B_sensitivity_heatmap}, and \ref{fig:Qwen3_32B_sensitivity_heatmap} show the normalized cross-lingual layer sensitivity heatmaps for LLaMA3-8B-Instruct, Qwen3-8B, and Qwen3-32B, respectively.

Overall, all three models exhibit a mixture of cross-lingually salient and language-specific sensitive layers. 
For LLaMA3-8B-Instruct, the shared pattern is particularly strong: a small number of layers, especially around Layer 9, show consistently high sensitivity across many languages. 
This suggests that LLaMA3-8B-Instruct contains strongly shared safety-sensitive layers that are activated across languages.

In contrast, the two Qwen3 models show more dispersed sensitivity patterns. 
Although some layers are repeatedly salient across languages, many high-sensitivity regions are more language-dependent. 
For example, Qwen3-8B exhibits sensitivity peaks across a wider range of middle and later layers, while Qwen3-32B shows more scattered language-specific peaks across its deeper layer stack. 
These results suggest that language-specific sensitivity is more pronounced in the Qwen3 models.

Together, these additional results support the main observation that sensitive layers are only partially shared across languages. 
While some layers are consistently salient, others exhibit language- or language-group-specific sensitivity, indicating that layer sensitivity depends on both model architecture and language.

\subsection{Selected Sensitive Layers}

Table~\ref{tab:selected-sensitive-layers} reports the language-shared and language-specific sensitive layers selected for each model.
The shared layer is selected by aggregating layer sensitivity across languages, while the language-specific layers are selected from each language's ranked sensitive layers after excluding the shared layer.
These layers are used in the scoring stage of \textsc{MMSafe}.

\subsection{Sensitivity to the Number of Language-Specific Layers}
\label{app:k_sensitivity}

We further study the effect of the number $k$ of language-specific
sensitive layers used by \textsc{MMSafe}. Using too few layers may
miss complementary safety-relevant signals, whereas including too
many layers may introduce weakly relevant or noisy representations.
We therefore compare $k\in\{1,2,3\}$ and report the summed number of
unsafe responses across evaluation settings; lower values indicate
better safety preservation.

\begin{table}[t]
\centering
\small
\begin{tabular}{lrrr}
\toprule
\textbf{Model} & $\mathbf{k=1}$ & $\mathbf{k=2}$ & $\mathbf{k=3}$ \\
\midrule
Qwen3-8B     & 4891 & \textbf{4576} & 4673 \\
LLaMA3.1-8B & 3894 & \textbf{3746} & 3937 \\
LLaMA3-8B    & \textbf{4810} & 4881 & 5031 \\
\bottomrule
\end{tabular}
\caption{Sensitivity to the number $k$ of language-specific sensitive
layers. We report summed unsafe-response counts across evaluation
settings; lower is better.}
\label{tab:k_sensitivity}
\end{table}

As shown in Table~\ref{tab:k_sensitivity}, $k=2$ achieves the best
performance on two of the three evaluated models and the best average
performance overall. While $k=1$ performs slightly better on
LLaMA3-8B, using two language-specific layers provides a more stable
default across models. Increasing $k$ to 3 does not yield further
improvements, suggesting that incorporating additional, less-sensitive
layers may introduce noisy signals. We therefore use $k=2$ as the
default rather than treating it as universally optimal for every model.

\begin{figure}[t]
  \centering
  \includegraphics[width=\linewidth]{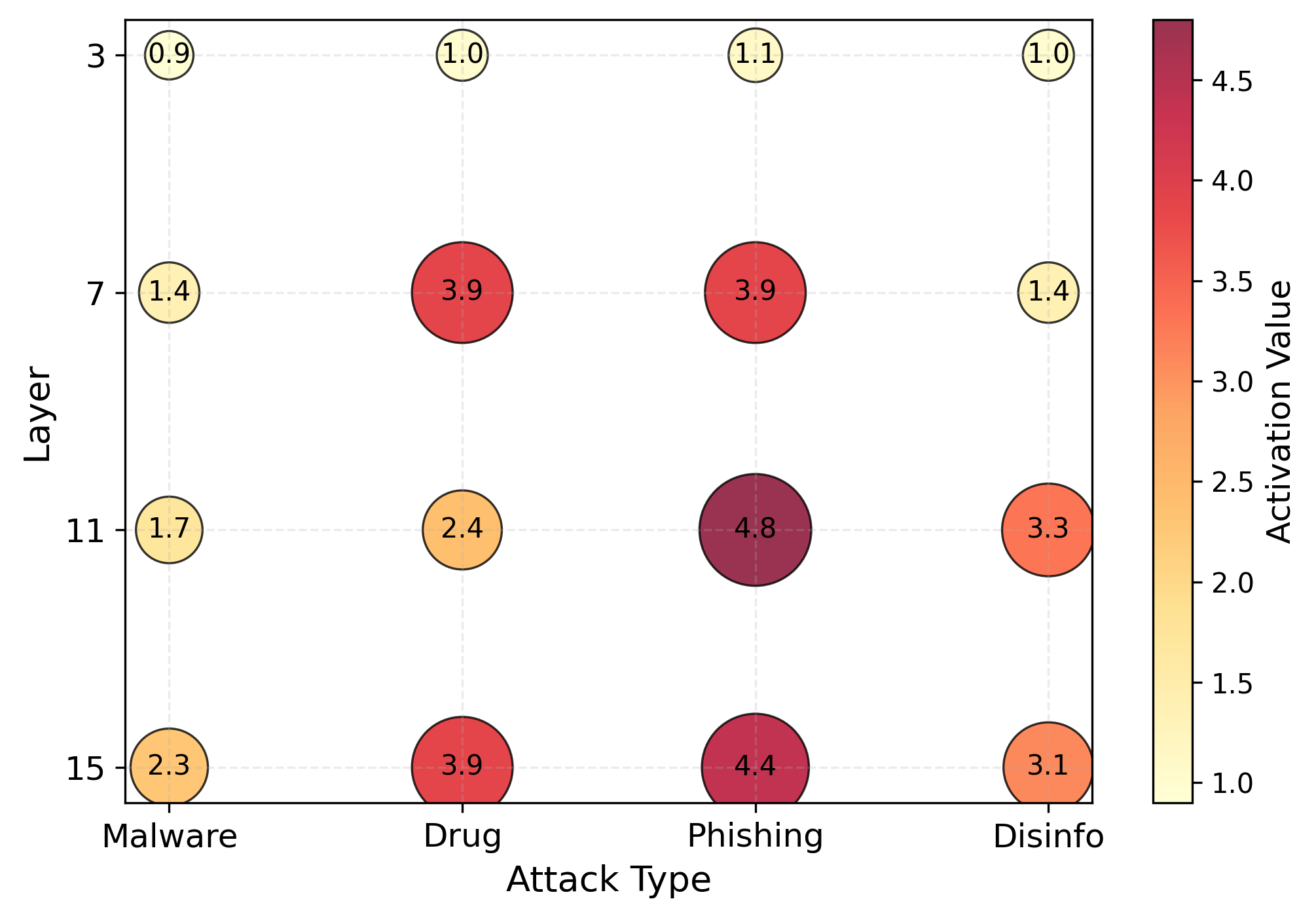}
  \vspace{-0.8em}
  \caption{
  Category-specific SAE activations across layers on HarmBench harmful queries using LLaMA3.1-8B-Instruct.
  Bubble size and color indicate activation strength.
  Different layers exhibit distinct activation patterns across harmful categories, suggesting that safety-relevant signals are semantically complementary across layers.
  }
  \vspace{-1.2em}
  \label{fig:sae-category-activation}
\end{figure}

\section{Safety Signals Analysis}
\label{app:safety-signals-analysis}

\subsection{Category-specific SAE Activations Across Layers}

To further examine whether different layers encode complementary safety information, we analyze SAE-interpreted feature activations on HarmBench harmful queries. 
Specifically, we consider four harmful categories, including malware, drug-related content, phishing, and disinformation, and measure the activation strength of harmful SAE features at Layers 3, 7, 11, and 15 of LLaMA3.1-8B-Instruct.

Figure~\ref{fig:sae-category-activation} shows that different layers exhibit distinct activation patterns across harmful categories. 
Layer 3 shows relatively weak and uniform activations, suggesting that it mainly captures lower-level lexical or local semantic cues. 
In contrast, Layer 7 is strongly activated by drug-related and phishing samples, while Layer 11 shows the strongest response to phishing and a higher response to disinformation. 
Layer 15 remains highly activated across multiple categories, especially malware, drug-related content, and phishing.

These results indicate that safety-relevant representations are not simply repeated across layers. 
Instead, different layers emphasize different harmful concepts and categories, providing complementary safety information. 
Together with the relative sensitivity analysis in the main text, this supports our claim that safety signals span multiple layers and are not fully captured by a single safety-sensitive layer.

\end{document}